\documentclass[12pt]{iopart}

\usepackage{graphicx}
\usepackage{dcolumn}
\usepackage{bm}

\include{epsf} 

\begin{document}
\title{Phylogeny and geometry of languages from normalized
Levenshtein distance}

\author{Maurizio Serva}
\address{Dipartimento di Matematica,
Universit\`a dell'Aquila, I-67010 L'Aquila, Italy}
\ead{serva@univaq.it}

\begin{abstract}

The idea that the distance among pairs of languages can be evaluated from lexical 
differences seems to have its roots in the work of the French
explorer Dumont D'Urville. 
He collected comparative words lists of various languages during his voyages 
aboard the Astrolabe from 1826 to 1829 and, in his work about the 
geographical division of the Pacific, he proposed a method to measure 
the degree of relation between languages. 

The method used by the modern lexicostatistics, developed by Morris
Swadesh in the 1950s, measures distances from the percentage of shared
cognates, which are words with a common historical origin. The weak
point of this method is that subjective judgment plays a relevant
role.

Recently, we have proposed a new automated method 
which is motivated by the analogy with genetics.  
The new approach avoids any subjectivity and results can be 
easily replicated by other scholars.
The distance between two languages is defined by considering a
renormalized Levenshtein distance between pair of words with the 
same meaning and averaging on the words contained in a list. 
The renormalization, which takes into account the length of the words,
plays a crucial role, and no sensible results can be found
without it.

In this paper we give a short review of our automated method and we 
illustrate it by considering the cluster of Malagasy dialects.
We show that it sheds new light on their kinship relation
and also that it furnishes a lot of new information concerning 
the modalities of the settlement of Madagascar.

\end{abstract}


\section*{Introduction}

The use of Swadesh lists \cite{Sw} in lexicostatistics is popular since half a century. 
They are lists of words associated to the same $M$ meanings,
(the original Swadesh choice was $M=200$) which concern the basic activities of humans. 
The choice is motivated by the fact that these terms, which are learned during
childhood, change very slowly over time.
Comparing the two lists corresponding to a pair of languages
it is possible to determine the percentage of shared {\it cognates}
which is a measure of their {\it lexical} distance.
Then,  provided that vocabularies change at a constant rate,
this lexical distance is roughly logarithmically proportional to 
the divergence time.
A recent example of the use of Swadesh lists and cognates counting to construct language 
trees are the studies of Gray and Atkinson \cite{GA} and Gray and Jordan \cite{GJ}.

Cognates are words with a common historical origin, 
their identification is often a matter of sensibility and personal knowledge. 
In fact, the task of counting the number of cognate 
words in the list is far from
trivial because cognates do not necessarily look similar.
Therefore, subjectivity plays a relevant role.
Furthermore,  results are often biased 
since it is easier for European or American scholars to find
out those cognates belonging to the western languages.
For instance, the Spanish word {\it leche} and the Greek 
word {\it gala} are cognates.  In fact, {\it leche} comes
from the  Latin {\it lac} with genitive form {\it lactis},
while the genitive form of {\it gala} is {\it galactos}.
Also the English {\it wheel} and the Hindi {\it cakra} are cognates. 
These two identifications were possible
because of our historical records, hardly they could have been
possible for languages, let's say, of Australia. 

The idea of measuring relationships among languages
using vocabulary is much older than lexicostatistics
and it seems to have its roots in the work of 
the French explorer Dumont D'Urville.
He collected comparative word lists of various
languages during his voyages aboard the Astrolabe 
from 1826 to 1829 and, in his work about 
the geographical division of the Pacific \cite{Urv},
he proposed a method to measure the degree of relation 
among languages.
He used a core vocabulary of 115 terms which, impressively, 
contains almost all the meanings of the 100-items lists of Swadesh.
Then, he assigned a distance from 0 to 1 to any pair of words with the 
same meaning  and finally he was able to determine the degree of
relation between any pair of languages.

Our automated method \cite{SP1,PS1} has some advantages: 
the first is that, at variance with previous approaches, 
it avoids subjectivity, the second is
that results can be replicated by other scholars assuming that the
database is the same, the third is that it is not requested a specific 
expertize in linguistic, and the last, but surely not the least, is
that it allows for a rapid comparison of a very large number of languages.
For any language we write down a Swadesh list, then 
we compare words with same meaning belonging to
different languages only considering orthographic differences.
This may appear reductive since words may look similar by chance,
while cognate words may have a completely different orthography,
but we will try to convince the reader that indeed this is a simpler,
more objective and more efficient choice with respect
to the traditional lexicostatistics approach.
This method is motivated by the analogy with genetics: 
the vocabulary has the role of DNA and the comparison is simply made by 
measuring the differences between the DNA of the two languages.

If a family of languages is considered, all the information is encoded 
in a matrix whose entries are the pairwise lexical distances, nevertheless,
this information is not manifest and it has to be extracted.  
The typical approach to this problem
is to transform the matrix information in a phylogenetic tree.
Nevertheless, the tree encodes only the information concerning the vertical
transmission between languages. 
Therefore, we also propose a complementary geometrical approach which 
transfers the matrix information into languages positions in a $n$-dimensional
euclidean space and which also takes into account the horizontal transmission.
The method is tested against the cluster of Malagasy dialects and it shows to
be able find out new important aspects of their internal organization and 
about their origin.

\section*{Lexical distances}

In order to describe our method, we start by our definition of lexical distance 
between two words, which is a variant of the Levenshtein distance \cite{Lev}. 
The Levenshtein distance is simply the minimum
number of  insertions, deletions, or substitutions of a
single character needed to transform one word into the other.
Our distance is obtained by a renormalization.

More precisely, given two words $\omega_1$ and $\omega_2$,
their distance $d(\omega_1, \omega_2)$ is defined as

\begin{equation}
d(\omega_1, \omega_2)= 
\frac{d_L(\omega_1, \omega_2)}{l(\omega_1, \omega_2)}
\label{wd}
\end{equation}
where $d_L(\omega_1, \omega_2)$ is their standard  Levenshtein distance
and $l(\omega_1, \omega_2)$ is the
number of characters of the longer of the two words
$\omega_1$ and $\omega_2$.
Therefore, the distance can take any value between 0 and 1.

The reason of the renormalization can be understood by
the following example.
Consider the case of two words with the same length
in which a single substitution 
transforms one word into the other.  
If they are short, let's say 2 characters, they are very different. 
On the contrary, if they are long, let's say 8 characters, 
it is reasonable to say that they are very similar. 
Without renormalization, their distance would be the same, equal to 1, 
regardless of their length. 
Instead, introducing the normalization factor, in the first 
case the distance is $\frac{1}{2}$,
whereas in the second, it is much smaller and equal to $\frac{1}{8}$. 

We use distance between pairs of words, as defined above, 
to construct the lexical distances of languages. 
For any language we prepare a list of words associated to the same 
$M$ meanings (we adopt the original Swadesh choice of $M=200$). 

Assume that the number of languages is $N$ and 
any language in the group is labeled by a Greek letter
(say $\alpha$) and any word of that language by 
$\alpha_i$ with $1 \leq i \leq M$. The same index $i$ 
corresponds to the same meaning in all languages i.e.,
two words $\alpha_i$ and $\beta_j$  
in the languages $\alpha$ and $\beta$ 
have the same meaning if $i=j$.

The lexical distance between two languages is then defined as 

\begin{equation}
D(\alpha, \beta)=  \frac{1}{M} \sum_{i=1}^M
d(\alpha_i, \beta_i)
\label{ld}
\end{equation}
It can be seen that $D(\alpha, \beta)$
is always in the interval [0,1] and obviously $D(\alpha, \alpha)=0$. 

The result of the analysis described above is a $N \times N$ upper 
triangular matrix whose entries are the $N(N-1)/2$ non-trivial 
lexical distances $D(\alpha, \beta)$ between all pairs of languages.

It is important to notice that although the matrix of distances encodes 
all the information concerning relationships among the $N$ languages,
this information is not manifest and it has to be extracted.  
The typical approach to this problem
is to transform the matrix in a phylogenetic tree.

Nevertheless, in this transformation, part of the information may be lost
because transfer among languages is not exclusively vertical
(as in mtDNA transmission from mother to child)
but it also can be horizontal (borrowings and, in extreme cases, creolization).

Another approach is the geometric one that results
from Structural Component Analysis (SCA) that we have recently proposed
\cite{Bl}.
This approach encodes the matrix information into the positions of the languages 
in a $n$-dimensional space. For large $n$ one 
recovers all the matrix content, but a low dimensionality, 
typically $n$=2 or $n$=3, is sufficient to grasp all the relevant information.

Other methods can be used for specific purposes 
(see \cite{PS2, PS3, PS4, PRS}) and other information can be extracted 
by the matrix combining different approaches and/or 
comparing with other information sources as, for example, 
the matrix of geographical distances between the 
homelands of languages \cite {SPVW}.

We would like also to mention that the method described here was later
used and developed by another large group of scholars \cite{Wic}. 
They placed the method at the core of an ambitious
project, the ASJP (The Automated Similarity Judgment Program) whose aim, 
in the words of its proponents, is "achieving a 
computerized lexicostatistics analysis of ideally all the world's languages".

\section*{Malagasy dialects}

We demonstrate now the method by applying it to the Malagasy 
dialects which are regional variants of the same language of Indonesian 
origin. Indeed, the nearest relative is Maanyan 
which is spoken by a Dayak community in Borneo \cite{Dahl, Ad1}.
A relevant contribution also comes from loanwords of other Indonesian languages
as Malay \cite{Ad2} and also African ones \cite{BW}.

The vocabulary was collected by the
author with the invaluable help of Joselin\`a Soafara N\'er\'e at 
the beginning of 2010. 
The dataset, which can be found in \cite{SP2}, 
consists of 200 words Swadesh lists for 23
dialects of Malagasy from all the areas of the island.
Since the number of dialects is $N$=23, the output of our method is a 
matrix with $N(N-1)/2=253$ non-trivial entries
representing all the possible lexical distances among dialects.

The information concerning the vertical transmission of vocabulary
from the proto-Malagasy to the contemporary dialects can be extracted by
a phylogenetic approach. There are various possible choices for the 
algorithm for the reconstruction of the family tree
(see \cite{SPVW} for a discussion of this point), 
we show in Fig. 1 the output of the Unweighted Pair 
Group Method Average (UPGMA).
The input data for the UPGMA tree are the pairwise 
separation times obtained from the lexical distances by means of a 
simple logarithmic rule (\cite{SP1, PS1}).
The absolute time-scale is calibrated by the results of the SCA analysis, 
which indicate a separation date A.D. 650 as it will be explained later.  
The phylogenetic tree in Fig. 1 interestingly shows a main 
partition of Malagasy dialects in two main branches (east-center-north 
and south-west) at variance with previous studies which gave a different
partitioning \cite{Ver} (indeed,
the results in \cite{Ver} coincide with ours if a correct
phylogeny is applied, see \cite{SPVW} for a discussion of this point.)   
Then, each of two branches splits, in turn, in two sub-branches
whose leaves are associated to different colors. 
In order to demonstrate the strict correspondence of this cladistic 
with the geography, we display a map of Madagascar (Fig. 2)
where the locations of the 23 dialects are indicated with the same 
colors of the leaves in Fig. 1.

Trees are ubiquitous in representations of languages taxonomies,
nevertheless, they fail to reveal all the information contained in the 
matrix of lexical distances. 
The reason is that the simple relation of ancestry, which is the
single principle behind a branching family tree model, 
cannot account for the complex interactions among dialects in {\it real time}.
Structural Component Analysis (SCA)
represents the relationships among different languages
as positions in a $n$-dimensional euclidean space (see \cite{Bl}
for a description of the method).
For a large number $n$ of dimensions all the information of the 
matrix is recovered, nevertheless, a low dimension ($n$=2 or $n$=3) 
is usually sufficient. 

In the case of Malagasy dialects, it happens that they belong to the 
same plane and therefore a dimension $n$= 2 is already enough. 
If one also considers an external language, as for example
Malay and Maanyan (which, as Malagasy, belong to the Austronesian family),
one finds that they stay on a different plane and, therefore,
the minimum dimension for a complete description is $n$=3.
In order to make this comparison it was necessary 
to compute the lexical distances between 
all the 23 Malagasy dialects and the two Indonesian languages.
The main source for Malay and Maanyan vocabularies was \cite{Gre}, implemented 
and corrected by the authors (our database can be consulted at \cite{SP2}). 
Any of the points in the 3-dimensional space can be individuated by its 
radial and angular coordinates. In Fig. 3 we plot the zenith angle
$\theta$ and azimuth angle $\varphi$ in the the two cases:
23 dialects + Maanyan and 23 dialects + Malay.
The angles clustering in Fig. 3 confirms the 4-partitions 
of Malagasy dialects and, in particular, it confirms
the relative isolation of the Antandroy variant (yellow)
which, nevertheless, belongs to the same plane of
the other Malagasy dialects, at variance with Malay and Maanyan.

It is worth to mention that the distribution of the dialects
along the radial direction is remarkably heterogeneous indicating that the 
rate of changes in the vocabulary was anything but stable over time. 
While the angular distribution gives informations concerning 
the proximity of languages and the clustering of the family, 
the radial distribution gives information about the date
in which the divergence of dialects started.
In fact, the variance of the radial distribution is
proportional to the time lag from the breaking of the unity of the 
proto-language \cite{Bl, SPVW}.
If the method is applied to our data, it gives a lag of about 1350 years.
Presumably, the beginning of divergence coincided with the colonization 
event, therefore, the variance gives the date 650 A.D. for the
landing of the Indonesian seafarers, a result which agrees with the 
date proposed by other scholars as Adelaar \cite{Ad1, Ad2}.

Finally, we plotted the distances of Malagasy dialects from Maanyan 
and Malay and we have shown them in Fig. 4. 
We observe  that some dialects 
(Antananarivo, Fianarantsoa, Manajary, Manakara)
have a smaller distance both from Maanyan and Malay.
This suggests a scenario according to which there was a migration 
on the highlands of Madagascar (Betsileo and Imerina regions)
shortly after the landing on the south-east coast (Manakara, Manajary).
The same indication for a south-east landing comes from the fact 
that linguistic diversity is higher in that region (see \cite{SPVW}). 
Furthermore, these findings are supported by information
concerning geography, in fact, an oceanic current from Indonesia links
the Sunda strait with the south-east coast of Madagascar \cite{SPVW}.

\section*{Discussion and outlook}

The method that we have presented has many advantages as rapidity,
objectivity and reproducibility, nevertheless, we think that the main 
progress is that it can be used in a kind of blind mode.
In this way, it is possible to avoid errors
which may arise from preconceptions which may influence the results 
of more traditional and subjective approaches.

We applied to Malagasy dialects, and we found a consistent new representation
of their phylogeny. We also found a date (650 A.D.) and a place
(south-east coast) for the founding event. 

Concerning Madagascar, a still unexplained mystery about the migration 
from Indonesia is that the closest language to Malagasy 
is Maanyan which is spoken by an ethnic group of Bornean Dayaks.  
The problem is that it is unlikely that the Dayaks 
headed the spectacular migration from Kalimantan to Madagascar, 
since they are forest dwellers with river navigation skills only.
We plan to investigate this mystery comparing, by our method,
Malagasy dialects with various Indonesian languages.

\section*{Acknowledgments}

The ideas and the methods presented here are mostly
the fruit of joint work with Filippo Petroni,
while important developments have been obtained in collaboration with
Philippe Blanchard, Dimitri Volchenkov and S\"oren Wichmann.
We warmly thank Daniel Gandolfo, Eric W. Holman, Armando G. M. Neves and
Michele Pasquini for helpful discussions.
We also thank Joselin\`a Soafara N\'er\'e for her invaluable help in the 
collection of the Malagasy word lists.

\newpage
\begin{figure}
\epsfysize=16.0truecm \epsfxsize=18.0truecm
\centerline{\epsffile{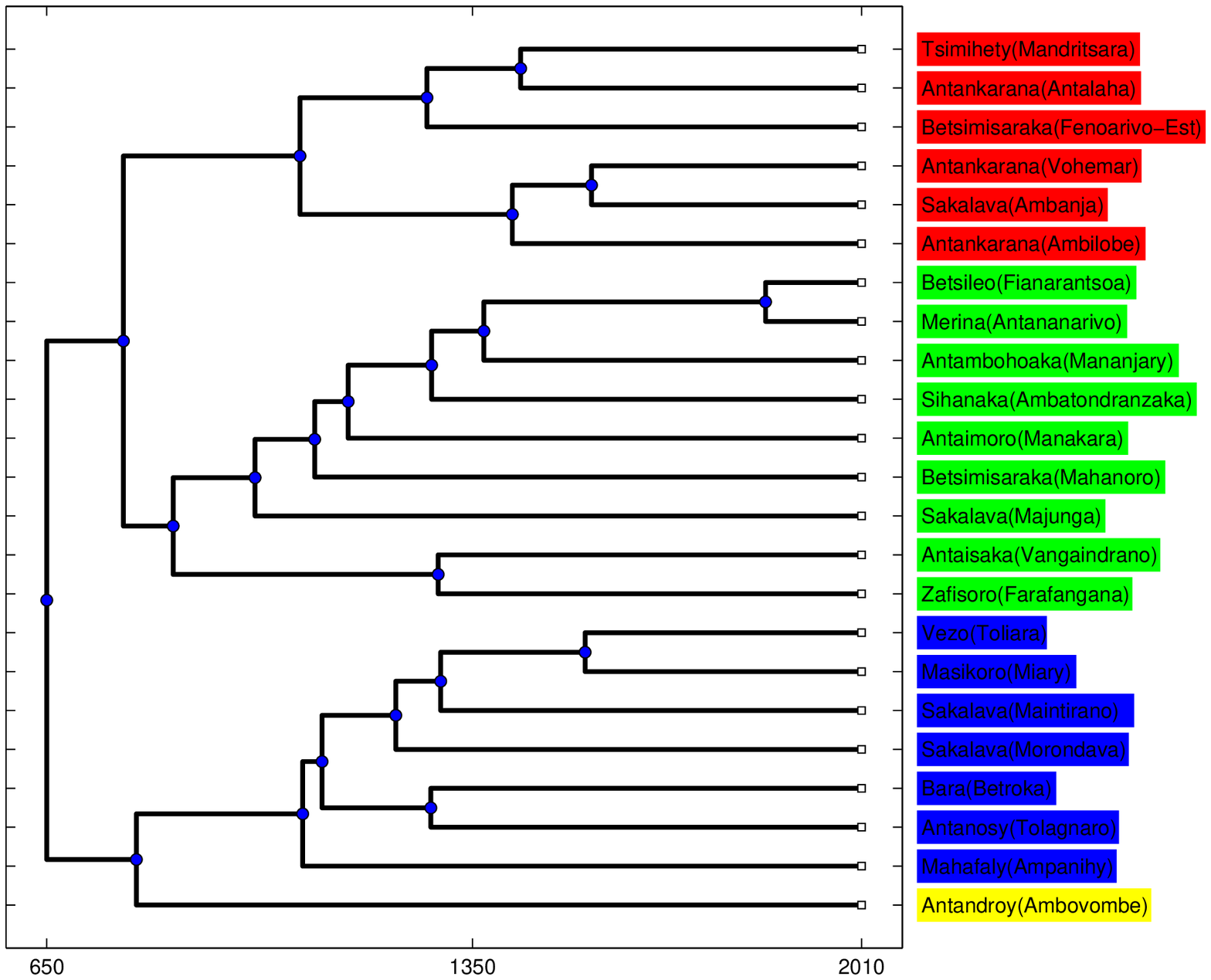}} 
\caption{Phylogenetic tree of 23 Malagasy dialects realized by
Unweighted Pair Group Method Average (UPGMA). 
The phylogenetic tree shows a main partition of the Malagasy dialects
into four main groups associated to different colors. 
The strict correspondence of this cladistic with the geography 
can be appreciate by comparison with Fig. 2.}
\label{fig1}
\end{figure}

\newpage
\begin{figure}
\epsfysize=20.0truecm \epsfxsize=26.0truecm
\centerline{\epsffile{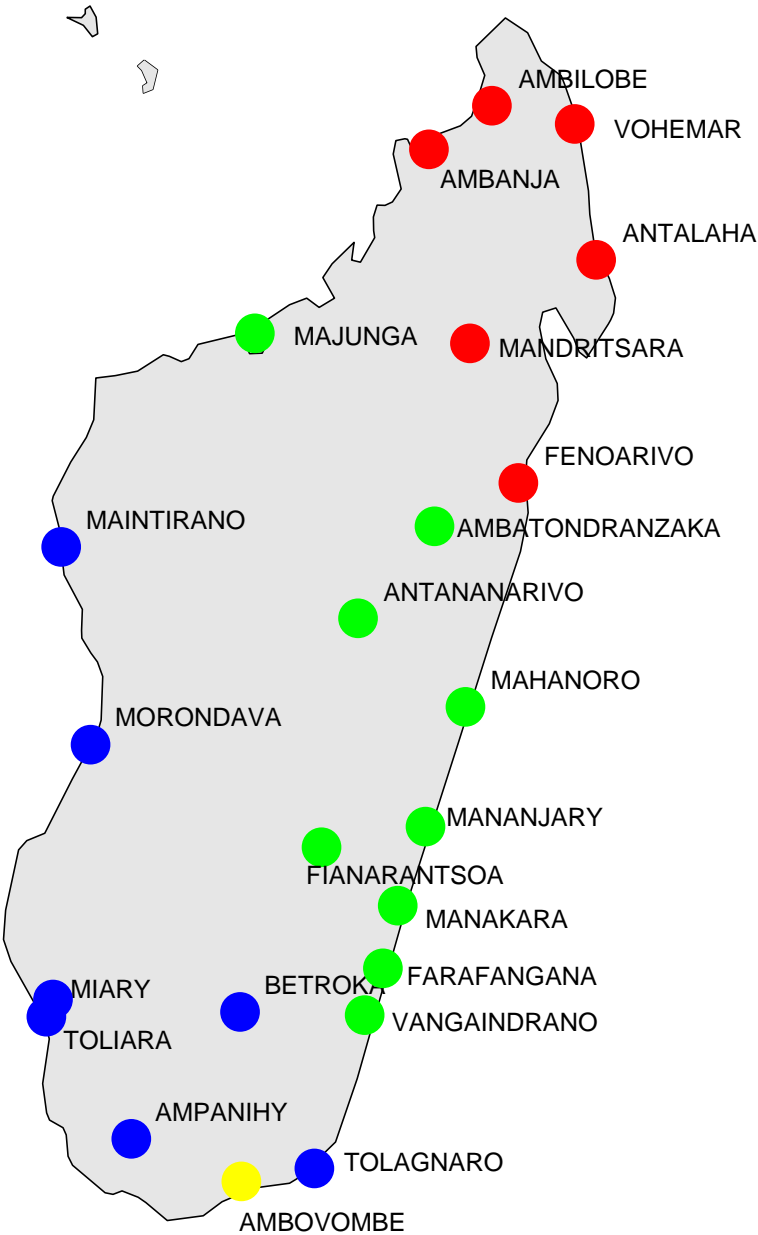}} 
\caption{Geography of Malagasy dialects.  The locations of the 23 dialects 
are indicated with the same colors of Fig. 1}
\label{fig2}
\end{figure}

\newpage
\begin{figure}
\epsfysize=16.0truecm \epsfxsize=16.0truecm
\centerline{\epsffile{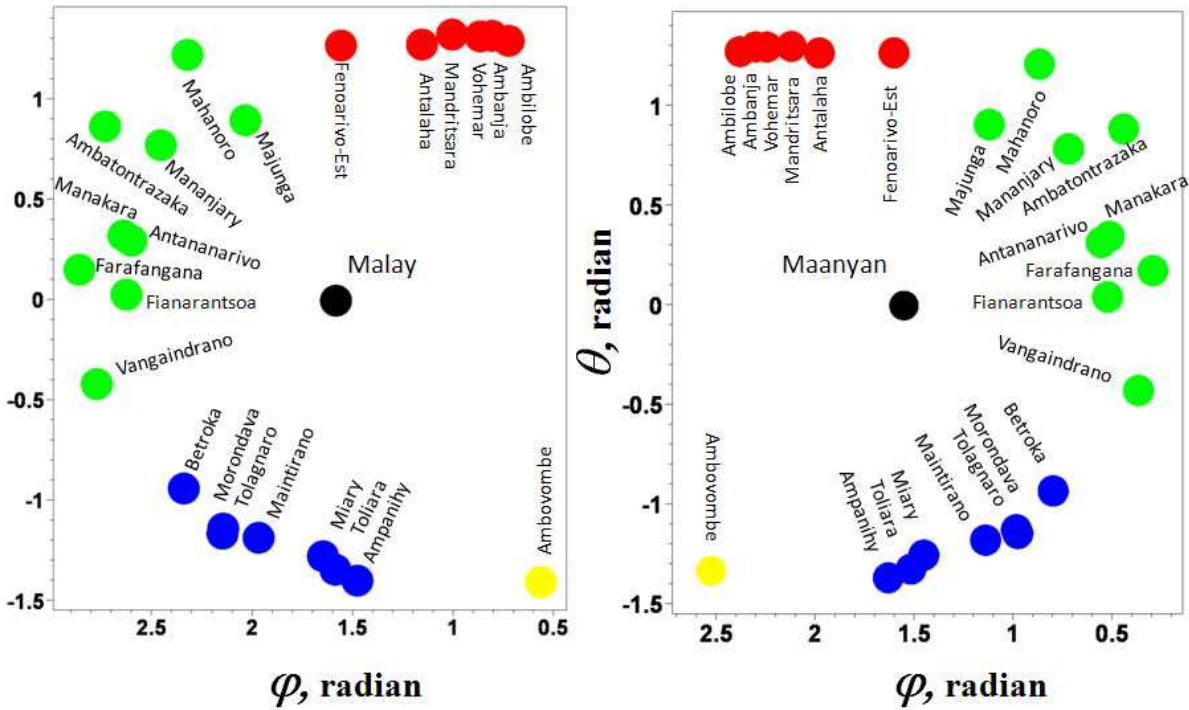}} 
\caption{Geometry of Malagasy dialects + Malay (left) and 
Malagasy dialects + Maanyan (right). Any dialect or language
is associated to a zenith angle $\theta$ and an azimuth angle $\varphi$. 
The radial component is not plotted. 
The dialects are indicated with the same colors of the map in Fig. 2.}
\label{fig3}
\end{figure}

\begin{figure}
\epsfysize=20.0truecm \epsfxsize=20.0truecm
\centerline{\epsffile{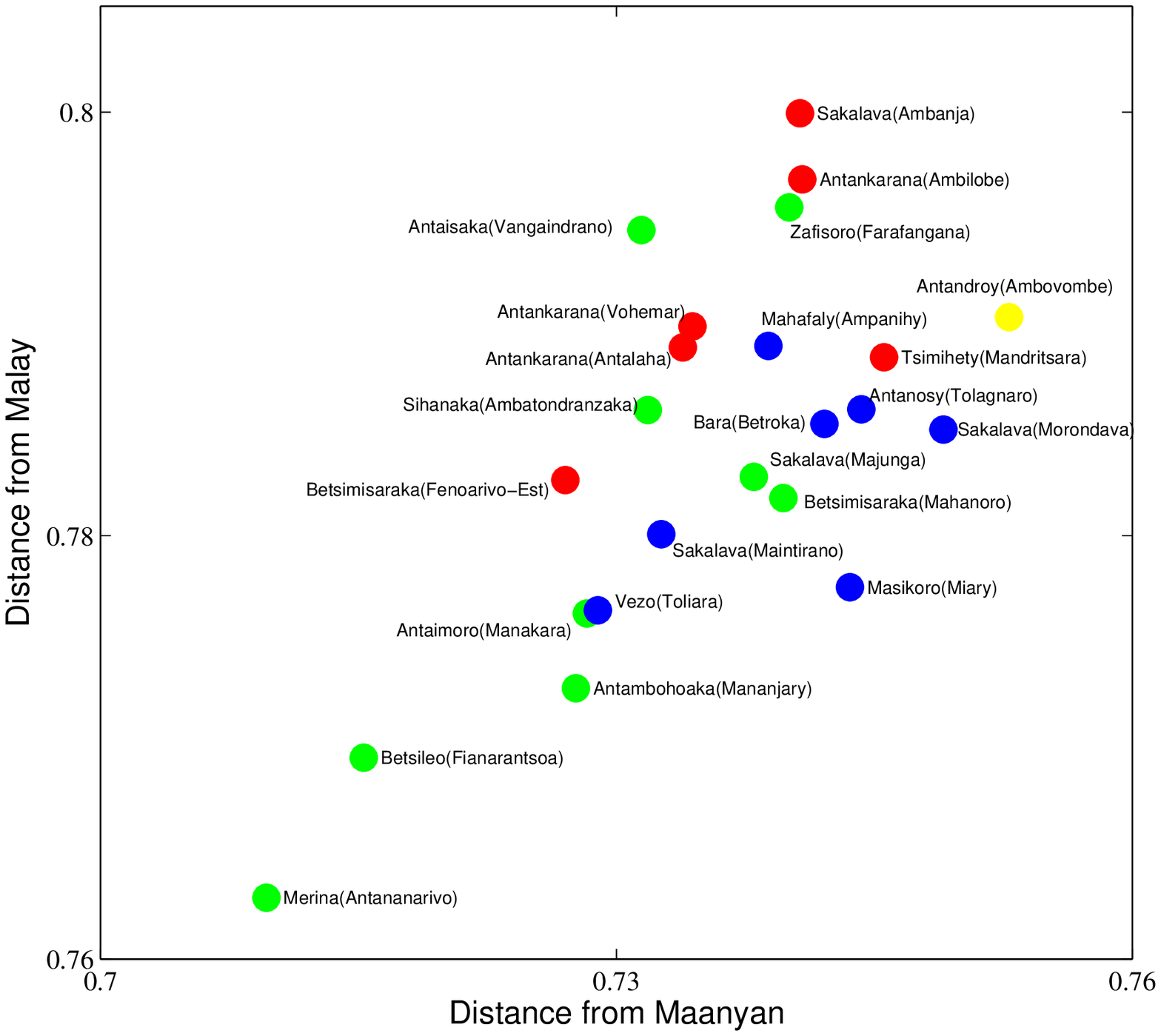}} 
\caption{Lexical distances of Malagasy dialects from Malay and Maanyan. 
The 23 dialects are indicated with the same colors of the map in Fig. 2.}
\label{fig4}
\end{figure}


\newpage
\section*{References}
\begin{thebibliography}{10}

\bibitem{Sw}
M. Swadesh, {\it Lexico-statistic dating of prehistoric ethnic contacts.} 
Proceedings American Philosophical Society, 
{\bf 96}, (1952), 452-463.

\bibitem{GA}
R. D. Gray and Q. D. Atkinson, 
{\it Language-tree divergence times support the Anatolian theory of 
Indo-European origin.}
Nature {\bf 426}, (2003), 435-439 

\bibitem{GJ}
R. D. Gray and F. M, Jordan,
{\it Language trees support the express-train sequence of 
Austronesian expansion.}
Nature {\bf 405}, (2000), 1052-1055.

\bibitem{Urv}
D. D'Urville,  {\it Sur les \^iles du Grand Oc\'ean}, 
Bulletin de la Soci\'et\'e de G\'eographie {\bf 17}, (1832), 1-21.

\bibitem{SP1}
M. Serva and F. Petroni,
{\it Indo-European languages tree by Levenshtein distance.}
EuroPhysics Letters {\bf 81}, (2008), 68005.

\bibitem{PS1}
F. Petroni and M. Serva,
{\it Languages distance and tree reconstruction.}
Journal of Statistical Mechanics: 
theory and experiment, (2008), P08012.  

\bibitem{Lev}
V. I. Levenshtein, 
{\it Binary codes capable of correcting deletions, insertions and reversals.} 
Soviet Physics Doklady {\bf 10}, 707 (1966).

\bibitem{Bl} 
Ph. Blanchard, F. Petroni, M. Serva and D. Volchenkov,
{\it Geometric representations of language taxonomies.}
Computer Speech and Language {\bf 25}, (2011), 679-699.

\bibitem{PS2}
F. Petroni and M. Serva,
{\it Lexical evolution rates by automated stability measure.}
Journal of Statistical Mechanics: 
Theory and Experiment, (2010), P03015, (10 pages).

\bibitem{PS3}
F. Petroni and M. Serva,
{\it Measures of lexical distance between languages.}
Physica A  {\bf 389}, (2010), 2280-2283.

\bibitem{PS4}
F. Petroni and M. Serva,
{\it Automated word stability and language phylogeny.}
Journal of Quantitative Linguistics {\bf 18}, (2011), 53-62.

\bibitem{PRS}
L. Prignano and M. Serva,
{\it Genealogical trees from genetic distances.}
European Physical Journal B  {\bf 69}, (2009), 455–463.

\bibitem{Wic}
S. Wichmann, A. M\"{u}ller, V. Velupillai, C. H. Brown, E. W. Holman, 
P. Brown, S. Sauppe, O. Belyaev, M. Urban, Z. Molochieva, A. Wett, 
D. Bakker, J-M. List, D. Egorov, R. Mailhammer, D. Beck and H. Geyer, 
{\it The ASJP Database}, (version 13, 2010).
http://email.eva.mpg.de/$\sim$wichmann/languages.htm.

\bibitem{SP2} 
M. Serva and F. Petroni, 
{\it Malagasy and related languages.} (Database updated in 2011).              
http://univaq.it/$\sim$serva/languages/languages.html. 

\bibitem{SPVW}
M. Serva, F. Petroni, D. Volchenkov and S. Wichmann,
{\it Malagasy dialects and the peopling of Madagascar.}
Journal of the Royal Society Interface (submitted). 
http://arxiv.org/pdf/1102.2180.

\bibitem{Gre}
S. J. Greenhill, R. Blust, and R.D. Gray,
{\it The Austronesian Basic Vocabulary Database.}
http://language.psy.auckland.ac.nz/austronesian).

\bibitem{Dahl}
O. C. Dahl,
{\it Malgache et Maanjan: une comparaison linguistique}. 
Oslo: Egede Instituttet (1951).

\bibitem{Ad1}
A. Adelaar,
{\it Borneo as a Cross-Roads for Comparative Austronesian Linguistics.}
In {\it The Austronesians in history.}
J. F. Bellwood, and D. Tryon editors, 75-95 (1995).
Australian National University, ANU E Press.

\bibitem{Ad2}
A. Adelaar,
{\it  Loanwords in Malagasy.} 
In {\it Loanwords in the World's Languages: A Comparative Handbook,} 
M. Haspelmath and U. Tadmor editors, 717-746 (2009). Berlin: De Gruyter Mouton.

\bibitem{BW}
R. M. Blench and M. Walsh,
{\it Faunal names in Malagasy: their etymologies and implications for
the prehistory of the East African Coast.}
In {\it Eleventh International Conference on
Austronesian Linguistics}
(11 ICAL), Aussois, France, (2009).

\bibitem{Ver}
P. V\'{e}rin, C.P. Kottak and P. Gorlin,
{\it The glottochronology of Malagasy speech communities.} 
Oceanic Linguistics {\bf 8}, 26-83 (1969).

\end{thebibliography}
\end{document}